\documentclass{llncs}
\usepackage{graphicx}
\usepackage{amsmath,amssymb} 
\usepackage{color}
\usepackage{array}
\usepackage{longtable}
\usepackage{tabu, soul}
\usepackage{xcolor, fancybox, framed, comment, xspace}

\definecolor{light-gray}{rgb}{0.8,0.8,0.8}
\definecolor{orange}{rgb}{1,0.7,0.2}
\definecolor{red}{HTML}{FF7F50}

\newcommand{\latinphrase}[1]{\textit{#1}}  

\newcommand{\ie}{\latinphrase{i.e.}\xspace}
\newcommand{\eg}{\latinphrase{e.g.}\xspace}    
\newenvironment{mynotes}[1][light-gray]
  {\def\FrameCommand{\fboxsep=\FrameSep\fcolorbox{#1}{#1}}%
    \MakeFramed {\advance\hsize-\width \FrameRestore}}
  {\endMakeFramed}
  
\newcommand{\jose}[1]{}

\begin{document}
\pagestyle{headings}
\mainmatter

\title{A discussion on the validation tests employed to compare human action recognition methods using the MSR Action3D dataset \\
\small{(updated: june 2015)}
}

\titlerunning{ }

\authorrunning{J.R. Padilla-L\'opez, A.A. Chaaraoui and F. Fl\'orez-Revuelta}

\author{Jos\'e Ram\'on Padilla L\'opez\inst{1} \and Alexandros Andr\'e Chaaraoui\inst{1} \and Francisco Fl\'orez Revuelta\inst{2} }

\institute{Department of Computer Technology, University of Alicante,\\
		   P.O. Box 99, E-03080 Alicante, Spain \\
\email{jpadilla@dtic.ua.es, alexandros@dtic.ua.es}
\and
Faculty of Science, Engineering and Computing, Kingston University,\\                  Penrhyn Road, KT1 2EE, Kingston upon Thames, United Kingdom \\
\email{F.Florez@kingston.ac.uk}}

\maketitle

\begin{abstract}
This paper aims to determine which is the best human action recognition method based on features extracted from RGB-D devices, such as the Microsoft Kinect. A review of all the papers that make reference to MSR Action3D, the most used dataset that includes depth information acquired from a RGB-D device, has been performed. We found that the validation method used by each work differs from the others. So, a direct comparison among works cannot be made. However, almost all the works present their results comparing them without taking into account this issue. Therefore, we present different rankings according to the methodology used for the validation in orden to clarify the existing confusion.

\end{abstract}

\section{Introduction}

\begin{table}[t]
\caption{State-of-the-art datasets for action recognition based on depth or skeletal features, sorted from more quoted to less quoted according to Google Scholar.}
\centering
\begin{tabu} to \linewidth{|X[3.4,l]|X[0.9,c]|X[0.9,c]|X[0.9,c]|X[1,c]|X[1,c]|X[0.8,c]|}
\hline
\textbf{\scriptsize Name} & \textbf{\scriptsize Actions} & \textbf{\scriptsize Actors} & \textbf{\scriptsize Times} & \textbf{\scriptsize Samples} & \textbf{\scriptsize Citations} & \textbf{\scriptsize Year} \\
\hline
MSR Action3D~\cite{Li2010} & 20 & 10 & 2 or 3 & 567 & 176 & 2010 \\
\hline
MSR DailyActivity3D~\cite{Wang2012} & 16 & 10 & 2 & 320 & 138 & 2012 \\
\hline
RGBD-HuDaAct~\cite{Ni2011} & 12 & 30 & 2 or 4 & 1189 & 86 & 2011 \\
\hline
CAD-60~\cite{Jaeyong2012} & 12 & 2+2 & - &	60 & 80 & 2012 \\
\hline
UTKinect Action~\cite{Xia2012} & 10 & 10 & 2 & - & 73 & 2012 \\
\hline
MSRC-12 KinectGesture~\cite{Fothergill2012} & 12 & 30 & - & 594 & 39 & 2012 \\
\hline
CAD-120~\cite{Koppula2013} & 10 & 2+2 & - & 120 & 33 & 2013 \\
\hline
MSR ActionPairs~\cite{Oreifej2013} & 6 & 10 & 3 & 180 & 29 & 2013 \\
\hline
MSR Gesture3D~\cite{Kurakin2012} & 12 & 10 & 2 or 3 & 336 & 25 & 2012 \\
\hline
LIRIS Human Activities~\cite{Wolf2012} & 10 & 21 & - & - & 24 & 2012 \\
\hline
Berkeley MHAD~\cite{Ofli2013} & 11 & 7+5 & 5 & $\sim660$ & 18 & 2013 \\
\hline
G3D~\cite{Bloom2013} & 20 & 10 & 3 & - & 11 & 2012 \\
\hline
ACT4 Dataset~\cite{Cheng2012} & 14 & 24 & \textgreater1 & 6844 & 9 & 2012 \\
\hline
UPCV Action~\cite{Theodorakopoulos2014} & 10 & 20 & - & - & 6 & 2014 \\
\hline
WorkoutSu-10 Gesture~\cite{Negin2013} & 10 & 15 & 10 & 1500 & 6 & 2013 \\
\hline
IAS-Lab Action~\cite{Munaro2013} & 15 & 12 & 3 & 540 & 3 & 2013 \\
\hline
Florence 3D Action~\cite{Seidenari2013} & 9 & 10 & 2 or 3 & 215 & 2 & 2012 \\
\hline
\end{tabu}
\label{tab:datasets}
\end{table}

In recent years, interest has grown on affordable devices (\eg \emph{Microsoft Kinect} or \emph{ASUS Xtion Pro}) that capture depth quite reliably. Such devices provide a depth image (D), along with an RGB image (thus RGB-D). A depth image can be further processed to obtain marker-less body pose estimation by means of a skeleton model consisting of a series of joints. Due to their low cost, high sample rate and capability to combine visual and depth information, these devices have become widespread in both research and commercial applications. Furthermore, their use has not been restricted to games, for which they were initially designed, but other applications where natural human-computer interaction is required.

These devices are widely used in the field of human action recognition (HAR), particularly in indoor scenarios for the recognition of activities of daily living. For research purposes, a variety of datasets for human action (or gesture) recognition have been recorded using RGB-D devices (see Table~\ref{tab:datasets}). The MSR Action3D dataset~\cite{Li2010} from Microsoft Research stands out as one of the most used in the literature, as many developed methods for action recognition have been validated with this dataset. Hence, it should be easy to determine the best human action recognition method in a straightforward way by comparing their success and processing rates. However, to the best of our knowledge, this is not possible at the moment as we found that almost all the works compare results obtained with different validation methods.

Therefore, this work aims to fill the existing gap in order to enable a fair comparison of the state of the art. We have reviewed 176 papers that make reference to the MSR Action3D dataset. Out of these 176 papers, 62 papers have been considered as they use the MSR Action3D dataset for the validation of the human action (or gesture) recognition methods proposed. They are classified according to the validation method and ranked based on their success rate. 

The remainder of this paper is organised as follows: Section~\ref{sec:msr-action3d} describes the MSR Action3D dataset employed by the reviewed works. In section~\ref{sec:instances}, an explanation of the inconsistencies found in the number of the used samples is given. Section~\ref{sec:validation-methods} presents the validation methods used in the reviewed papers and provides a classification of each work according to this. Finally, section~\ref{sec:conclusion} presents some conclusions and recommendations for the future.

\section{MSR Action3D dataset}
\label{sec:msr-action3d}
The MSR Action3D dataset \cite{Li2010} contains 20 different actions, performed by 10 different subjects with up to 3 different repetitions. This makes a total of 567 sequences and each one includes depth and skeleton joints. However 10 sequences are not valid in this dataset because the skeletons were either missing or wrong, as explained by the authors\footnote{MSR Action Recognition Datasets and Codes, http://research.microsoft.com/en-us/um/people/zliu/actionrecorsrc/default.htm (last access: 06/26/2014)}. The authors divided the dataset in three subsets of 8 gestures each, as shown in Table~\ref{tab:subsets}. Most of the papers working with this dataset have also used them. This was due to the high computational cost of dealing with the overall dataset. The AS1 and AS2 subsets were intended to group actions with similar movement, while AS3 was intended to group complex actions together. 

\begin{table}[t]
\caption{Actions in each of the MSR Action3D subsets.}
\centering  
\begin{tabular}{|c | l | c | l | c | l |} 
\hline
\multicolumn{2}{|c|}{\textbf{AS1}} & \multicolumn{2}{c}{\textbf{AS2}} & \multicolumn{2}{|c|}{\textbf{AS3}}\\
\hline
Label & Action name & Label & Action name & Label & Action name\\
\hline
a02 & Horizontal arm wave & a01 & High arm wave	& a06 & High throw \\
a03 & Hammer & a04 & Hand catch	& a14 & Forward kick \\
a05 & Forward punch & a07 & Draw cross	& a15 & Side-kick \\
a06 & High throw & a08 & Draw tick	& a16 & Jogging \\
a10 & Hand clap & a09 & Draw circle	& a17 & Tennis swing \\
a13 & Bend & a11 & Two-hand wave	& a18 & Tennis serve \\
a18 & Tennis serve & a14 & Forward kick	& a19 & Golf swing \\
a20 & Pick-up and throw & a12 & Side-boxing		& a20 & Pick-up and throw \\
\hline 
\end{tabular}
\label{tab:subsets} 
\end{table}

\section{How many samples are used for testing?}
\label{sec:instances}
Despite of the fact that the MSR Action3D dataset is made up of 567 sequences, the number of instances used in some works is unclear~\cite{Azary2013,Gao2014,Shen2014}. There is a lot of confusion concerning this topic.

As far as we know, the authors of the dataset firstly described it as made up of twenty actions, where each one was performed by seven subjects for three times~\cite{Li2010}. However, actions are performed by ten subjects with up to three repetitions as described in the previous section. Many works have compared their results with Li et al. and most of them used ten subjects~\cite{Xia2012,Wang2013,Yang2014}. In other words, they may have used a higher number of instances than the work they aim to compare to. Wang et al.~\cite{Wang2012} described the dataset as made up of 402 sequences. For the sake of clarity, this mistake is advertised at the dataset web page\footnote{A list of the used sequences is also provided in the website}. The authors explain that 10 sequences out of the 567 are not used because a number of skeletons are either missing or too erroneous. So, the dataset is eventually composed of 557 sequences. However, it is curious to see how recent works~\cite{Gao2014,Shen2014,Yuan2014} still mention that the dataset is composed of 402 sequences and directly compare their results with the state-of-the-art papers that use other number of instances. Furthermore, other authors have intentionally used a subset of the whole dataset, \eg 17 actions, 8 subjects and 3 repetitions (408 samples). Due to this, the AS1, AS2 and AS3 subsets are composed of different actions too, thereby they compare their results with works that use a different number of instances.

As a consequence, it is very difficult to confirm whether these works use 402, 557 or 567 samples as we are not sure whether the authors are aware of these key aspects concerning the dataset, or if those are only naive text mistakes. Moreover, the missing information concerning the number of instances prevents to make a fair comparison between different methods.

\section{Which is the validation method used?}
\label{sec:validation-methods}
Regarding the experimentation method used by many authors working with the MSR Action3D dataset, it is worth to mention that there is a lack of agreement. In the paper by Li et al.~\cite{Li2010} where the dataset was firstly presented, three tests are performed: 1/3, 2/3 and cross-subject test. In the first two tests, 1/3 and 2/3 of the instances are respectively used as training samples and the rest as testing samples. In the third test, half of the subjects are used for training and the remainder for testing. However, it is not described which instances or subjects are actually used in each partition of the dataset.

Given that information is missing, we could assume that the 1/3 means to split the dataset using the first repetition of each action performed by each subject as training, and to use the remainder for testing. The same could be assumed for the 2/3. However, if we only consider instances as a whole, we can split the dataset in a different way. For instance, the dataset can be split using 1/3 (or 2/3) of all the instances for training. The same is true for the cross-subject test. It is not stated which instances are used. Any half of all the subjects can be used for training, \eg 1, 2, 3, 9 and 10; and the remainder for testing, \ie 4, 5, 6, 7 and 8. Given that it is not clear which instances are used, each researcher is free to interpret anything, thereby comparing different methods where a distinct methodology has been used for the experimentation. However, this is not desirable to compare and decide which method performs better.

In the cross-subject test employed by Li et al.~\cite{Li2010} the actual samples of subjects 1, 3, 5, 7 and 9 are used for training, whereas actors 2, 4, 6, 8 and 10 are used for validation. This test is followed by many authors as shown in Table~\ref{tab:li-cross-subject-test}. While some authors use the mentioned settings for their training and validation sets, other authors use subjects 1-5 for training and 6-10 for validation (see Table~\ref{tab:1-5-train-6-10-test}). Regardless of the used setup, most of the works state that they follow the same settings as Li et al. but do not provide a description of such a setup. Due to this, we assume that they follow the same validation than Li et al., so Table~\ref{tab:li-cross-subject-test} and Table~\ref{tab:1-5-train-6-10-test} can even have classification mistakes. Anyway, a fair comparison cannot be performed. Indeed, when it is sure that the same setup has been used, sometimes results only show an accuracy score and the authors do not give an explanation of what it represents, \ie the average of the AS1, AS2 and AS3 tests, or the overall accuracy of using the whole dataset (20 actions).

\tabulinesep = 0.9mm
\begin{longtabu} to \linewidth{|>{\scriptsize}X[5.7,m]|>{\scriptsize}X[1,c]|>{\scriptsize}X[1,c]|>{\scriptsize}X[1,c]|>{\scriptsize}X[1,c]|>{\scriptsize}X[1,c]|>{\scriptsize}X[1,c]|}
\caption{Li et al.'s cross-subject test (1-3-5-7-9 training, 2-4-6-8-10 test). The first 12 methods explicitly describe both the training and validation sets. Results are ordered by the average result for the AS1, AS2, and AS3 subsets; and then by the results for the whole dataset.}
\label{tab:li-cross-subject-test} \\
\hline
\textbf{Method} & \textbf{Year} & \textbf{AS1} & \textbf{AS2} & \textbf{AS3} & \textbf{Avg.} & \textbf{All} \\
\hline
\endfirsthead
\multicolumn{7}{c}%
{\tablename\ \thetable\ -- \textit{Continued from previous page}} \\
\hline
\textbf{Method} & \textbf{Year} & \textbf{AS1} & \textbf{AS2} & \textbf{AS3} & \textbf{Avg.} & \textbf{All} \\
\hline
\endhead
\hline
\multicolumn{7}{c}{\textit{Continued on next page}} \\
\endfoot
\hline
\endlastfoot
Deep Convolutional Neural Networks for Action Recognition Using Depth Map Sequences, Wang et al. \cite{Wang2015} & 2015 & - & - & - & - & ¿100? \\
\hline
Efficient Pose-Based Action Recognition, Eweiwi et al. \cite{Eweiwi2015} & 2015 & - & - & - & - & 92.3? \\
\hline
Action Recognition from Depth Sequences Using Depth Motion Maps-based Local Binary Patterns,
Chen et al. \cite{Chen2015} & 2015 & 98.1 & 92 & 94.6 & 94.9 & 91.94 \\
\hline
HOPC: Histogram of Oriented Principal Components of 3D Pointclouds for Action Recognition, Rahmani et al. \cite{Rahmani2014eccv} & 2014 & - & - & - & - & 91.64? \\
\hline
Fusion of Skeletal and Silhouette-Based Features for Human Action Recognition with RGB-D Devices, Chaaraoui et al. \cite{Chaaraoui2013} & 2013 & 92.38  & 86.61  & 96.4  & 91.8  & - \\
\hline
Real-time human action recognition based on depth motion maps, Chen et al. \cite{Chen2013} & 2013 & 96.2  & 83.2  & 92  & 90.47  & - \\
\hline
Skeletal Quads: Human Action Recognition Using Joint Quadruples, Evangelidis et al. \cite{Evangelidis2014} & 2014
& 88.39  & 86.61  & 94.59  & 89.86  & - \\
\hline
Random Occupancy Patterns, Wang et al. \cite{Wang2014}
& 2014 & - & - & - & 86.50? & - \\
\hline
Action recognition based on a bag of 3d points, Li et al. \cite{Li2010}
& 2010 & 72.9  & 71.9 & 79.2 & 74.67 & - \\
\hline
Learning Maximum Margin Temporal Warping for Action Recognition, Wang and Wu \cite{Wang2013}
& 2013 & - & - & - & - & 92.7? \\
\hline
Learning Actionlet Ensemble for 3D Human Action Recognition, Yuan et al. \cite{Yuan2014}
& 2014 & - & - & - & - & 88.2? \\
\hline
Mining actionlet ensemble for action recognition with depth cameras, Wang et al. \cite{Wang2012}
& 2012 & - & - & - & - & 88.2? \\
\hline
\hline
\hline
Group Sparsity and Geometry Constrained Dictionary Learning for Action Recognition from Depth Maps, Luo et al. \cite{Luo2013}
& 2013 & 97.2  & 95.5  & 99.1  & 97.26  & 96.7  \\
\hline
Fusing Spatiotemporal Features and Joints for 3D Action Recognition, Zhu et al. \cite{Zhu2013}
& 2013 & - & - & - & 94.3 & - \\
\hline
Human Action Recognition by Mining Discriminative Segment with Novel Skeleton Joint Feature, Zou et al. \cite{Zou2013}
& 2013 & - & - & - & 94.0? & \\
\hline
Pose-based human action recognition via sparse representation in dissimilarity space , Theodorakopoulos et al. \cite{Theodorakopoulos2014}
& 2014 & 91.23  & 90.09  & 99.5  & 93.61  & - \\
\hline
Action recognition on motion capture data using a dynemes and forward differences representation, Kapsouras and Nikolaidis \cite{Kapsouras2014} & 2014
& - & - & - & 93.6  & 91.4  \\
\hline
Super Normal Vector for Activity Recognition Using Depth Sequences, Yang and Tian \cite{Yang2014b} & 2014
& - & - & - & 93.09? & - \\
\hline
Action Recognition Using Ensemble Weighted Multi-Instance Learning, Chen et al. \cite{Chen2014} & 2014
& -  & - & - & 92? & -  \\
\hline
Histogram of Oriented Displacements (HOD): Describing Trajectories of Human Joints for Action Recognition, Gowayyed et al. \cite{Gowayyed2013} & 2013 & 92.39  & 90.18  & 91.43  & 91.26  & - \\
\hline
DMM-Pyramid Based Deep Architectures for Action Recognition with Depth Cameras, Yand and Yang \cite{YangYang2015} & 2015 & - & - & - & 91.21? & - \\
\hline
Human Action Recognition Using a Temporal Hierarchy of Covariance Descriptors on 3D Joint Locations, Hussein et al. \cite{Hussein2013} & 2013
& 88.04  & 89.29  & 94.29  & 90.53  & - \\
\hline
Body Surface Context: A New Robust Feature for Action Recognition From Depth Videos, Song et al. \cite{Song2014} & 2014 
& - & - & - & 90.36? & - \\
\hline
An Approach to Pose-Based Action Recognition, Wang et al. \cite{Wang2013b} & 2013
& - & - & - & 90.22 & - \\
\hline
Human Action Recognition Via Multi-modality Information, Gao et al. \cite{Gao2014} & 2014
& 92  & 85  & 93  & 90  & - \\
\hline
Human Behavior Recognition Based on Axonometric Projections and PHOG Feature, Shen et al. \cite{Shen2014} & 2014 & 90.6  & 81.4  & 94.6  & 88.87  & - \\
\hline
On the improvement of human action recognition from depth map sequences using Space–Time Occupancy Patterns, Vieira et al. \cite{Vieira2014} & 2014 & 91.7  & 72.2  & 98.6  & 87.5  & 81.55  \\
\hline
STOP: Space-Time Occupancy Patterns for 3D Action Recognition from Depth Map Sequences, Vieira et al. \cite{Vieira2012} & 2012 & 84.7  & 81.3  & 88.4  & 84.8  & - \\
\hline
Effective 3D action recognition using EigenJoints, Yang And Tian \cite{Yang2014} & 2014
& - & - & - & 83.3? & - \\
\hline
Home Monitoring Musculo-skeletal Disorders with a Single 3D Sensor, Wang et al. \cite{Wang2013c} & 2013 & - & - & - & 81.9? & - \\
\hline
Online Human Gesture Recognition from Motion Data Streams, Zhao et al. \cite{Zhao2013} & 2013 & - & - & - & 81.7? & - \\
\hline
Effective approaches in human action recognition, Li et al. \cite{Li2013} & 2013
& - & - & - & 81.5 or 91.5? & - \\
\hline
Gesture recognition from depth images using motion and shape features, Qin et al. \cite{Qin2013} & 2013 & 81  & 79  & 82  & 80.66 & - \\
\hline
Human activity recognition using multi-features and multiple kernel learning, Althloothi et al. \cite{Althloothi2014} & 2014 & 74.3  & 76.8  & 86.7  & 79.27  & - \\
\hline
View invariant human action recognition using histograms of 3D joints, Xia et al. \cite{Xia2012} & 2012 & 87.98  & 85.48  & 63.46  & 78.97  & - \\
\hline
Three Dimensional Motion Trail Model for Gesture Recognition, Liang and Zheng \cite{Liang2013} & 2013 & 73.7  & 81.5  & 81.6 & 78.93 & -\\
\hline
Attractor-Shape for Dynamical Analysis of Human Movement: Applications in Stroke Rehabilitation and Action Recognition, Venkataraman et al. \cite{Venkataraman2013} & 2013 & 77.5  & 63.1  & 87  & 75.87  & - \\
\hline
Exploring the Trade-off Between Accuracy and Observational Latency in Action Recognition, Ellis et al. \cite{Ellis2013} & 2013 & - & - & - & 65.7? & - \\
\hline
Robust 3D Action Recognition with Random Occupancy Patterns, Wang et al. \cite{Wang2012b} & 2012 & - & - & - & - & 86.5? \\
\hline
\end{longtabu}

\normalsize

\begin{longtabu} to \linewidth{|>{\scriptsize}X[5.7,m]|>{\scriptsize}X[1,c]|>{\scriptsize}X[1,c]|>{\scriptsize}X[1,c]|>{\scriptsize}X[1,c]|>{\scriptsize}X[1,c]|>{\scriptsize}X[1,c]|}
\caption{Cross-subject test (1-5 training, 6-10 test). The first seven methods explicitly describe both the training and validation sets.}
\label{tab:1-5-train-6-10-test} \\
\hline
\textbf{Method} & \textbf{Year} & \textbf{AS1} & \textbf{AS2} & \textbf{AS3} & \textbf{Avg.} & \textbf{All} \\
\hline
\endfirsthead
\multicolumn{7}{c}%
{\tablename\ \thetable\ -- \textit{Continued from previous page}} \\
\hline
\textbf{Method} & \textbf{Year} & \textbf{AS1} & \textbf{AS2} & \textbf{AS3} & \textbf{Avg.} & \textbf{All} \\
\hline
\endhead
\hline
\multicolumn{7}{c}{\textit{Continued on next page}} \\
\endfoot
\hline
\endlastfoot
Sparse spatio-temporal representation of joint shape-motion cues for human action recognition in depth sequences, Tran and Ly \cite{Tran2013b} & 2013
& - & -& - & 91.92? & - \\
\hline
The Moving Pose: An Efficient 3D Kinematics Descriptor for Low-Latency Action Recognition and Detection, Zanfir et al. \cite{Zanfir2013} & 2013
& - & - & - & 91.7? & - \\
\hline
An effective fusion scheme of spatio-temporal features for human action recognition in RGB-D video, Tran and Ly \cite{Tran2013} & 2013
& - & - & - & 88.89? & - \\
\hline
Real Time Action Recognition Using Histograms of Depth Gradients and Random Decision Forests, Rahmani et al. \cite{Rahmani2014wacv} & 2014
& - & - & - & 88.8? & - \\
\hline 
Iterative temporal learning and prediction with the sparse online echo state gaussian process, Soh and Demiris \cite{Soh2012} & 2012
& 80.6 & 74.9  & 87.1  & 80.87  & - \\
\hline 
Joint Angles Similarities and HOG2 for Action Recognition, Ohn-Bar and Trivedi \cite{Ohn-Bar2013} & 2013
& - & - & - & - & 94.84 \\
\hline
HON4D: Histogram of Oriented 4D Normals for Activity Recognition from Depth Sequences, Oreifej and Liu \cite{Oreifej2013} & 2013
& - & - & - & - & 88.89?\\
\hline
\hline
\hline
Spatio-temporal feature extraction and representation for RGB-D human action recognition, Luo et al. \cite{Luo2014} & 2014 & 96.1  & 90.8  & 98.33  & 95.08  & 93.83?\\
\hline
Action Classification with Locality-constrained Linear Coding, Rahmani et al. \cite{Rahmani2014icpr} & 2014 & -  & - & - & 90.9? & - \\
\hline
Spatio-temporal Depth Cuboid Similarity Feature for Activity Recognition Using Depth Camera, Xia and Aggarwal \cite{Xia2013} & 2013 & - & - & - & 89.3? & - \\
\hline
Optimal Joint Selection for Skeletal Data from RGB-D Devices Using a Genetic Algorithm, Climent et al. \cite{Climent2013} & 2013 & -  & - & - & - & 71.1 \\
\hline
\end{longtabu}

Due to all this confusion about how to split the dataset in two sets for training and validation, some authors randomly choose half of the subjects for the training set, and select the rest of the subjects for the validation set. As in 2-fold cross validation, they repeat the test using the previous validation set as the training set and vice versa. In this case, the final result is the average of both tests (see Table~\ref{tab:2-fold-cross-validation}). In other works, instead of performing a 2-fold cross validation, some authors randomly select the two sets and repeat the experiment several times. For example, Miranda et al.~\cite{Miranda2012} perform a random selection of half of the actors as training set and the other half as validation set. This is repeated 10 times and the final result is the average of the results of each run. Other authors repeat the test 100 times instead of 10~\cite{Çeliktutan2013,Çeliktutan2014}, and even 200 times~\cite{Chen2013} (see Table~\ref{tab:miranda-test}). However, although the tests are repeated many times, all the possible splits are not considered, \ie all the possible combinations (252 tests) of using 5 subjects for training and the remaining ones for testing. Only a few works perform this test. In Table~\ref{tab:miranda-test} these works have been included with the 252 number in the third column. This indicates that they performed the test with all the possible combinations.

\tabulinesep = 0.6mm
\begin{table}
\caption{2-fold cross-validation test} 
\begin{tabu} to \linewidth{|>{\scriptsize}X[5.7,m]|>{\scriptsize}X[1,c]|>{\scriptsize}X[1,c]|>{\scriptsize}X[1,c]|>{\scriptsize}X[1,c]|>{\scriptsize}X[1,c]|>{\scriptsize}X[1,c]|}
\hline
\textbf{Method} & \textbf{Year} & \textbf{AS1} & \textbf{AS2} & \textbf{AS3} & \textbf{Avg.} & \textbf{All} \\
\hline
Evolutionary joint selection to improve human action recognition with RGB-D devices, Chaaraoui et al. \cite{Chaaraoui2014} & 2014 & 91.59  & 90.83  & 97.28  & 93.23  & - \\
\hline
Combining unsupervised learning and discrimination for 3D action recognition, Chen et al. \cite{ChenG2015} & 2015 & - & - & - & 91,4 & - \\
\hline
3D Action Classification Using Sparse Spatio-temporal Feature Representations, Azary and Savakis \cite{Azary2012} & 2012 & 77.66  & 73.17  & 91.58  & 80.8  & 63.23 \\
\hline 
\end{tabu} 
\label{tab:2-fold-cross-validation} 
\end{table}
 
\begin{table}
\caption{Miranda et al.'s test: Random selection of training and test sets repeated a number of times.} 
\begin{tabu} to \linewidth{|>{\scriptsize}X[5.7,m]|>{\scriptsize}X[1,c]|>{\scriptsize}X[1,c]|>{\scriptsize}X[1,c]|>{\scriptsize}X[1,c]|>{\scriptsize}X[1,c]|>{\scriptsize}X[1,c]|>{\scriptsize}X[1,c]|}
\hline
\textbf{Method} & \textbf{Year} & \textbf{Tests} & \textbf{AS1} & \textbf{AS2} & \textbf{AS3} & \textbf{Avg.} & \textbf{All} \\
\hline
Efficient Pose-Based Action Recognition, Eweiwi et al. \cite{Eweiwi2015} & 2015 & 252 & - & - & - & - & 88.38? \\
\hline
HOPC: Histogram of Oriented Principal Components of 3D Pointclouds for Action Recognition, Rahmani et al. \cite{Rahmani2014eccv} & 2014 & 252& - & - & - & - & 86.49? \\
\hline
Sparse spatio-temporal representation of joint shape-motion cues for human action recognition in depth sequences, Tran and Ly \cite{Tran2013b} & 2013 & 252
& - & - & - & 84.54? & -\\
\hline 
Real Time Action Recognition Using Histograms of Depth Gradients and Random Decision Forests, Rahmani et al. \cite{Rahmani2014wacv} & 2014 & 252
& - & - & - & - & 82.7? \\
\hline
HON4D: Histogram of Oriented 4D Normals for Activity Recognition from Depth Sequences, Oreifej and Liu \cite{Oreifej2013} & 2013 & 252
& - & - & - & - & 82.15? \\
\hline 
Real-time human action recognition based on depth motion maps, Chen et al. \cite{Chen2013} & 2013 & 200 & 90.1  & 90.6  & 97.6  & 92.77  & - \\
\hline 
Fast Exact Hyper-graph Matching with Dynamic Programming for Spatio-temporal Data, \c{C}eliktutan et al. \cite{Çeliktutan2014} & 2014 & 100 & 84.5  & 85  & 72.2  & 80.57 & - \\
\hline 
Graph-based Analysis of Physical Exercise Actions, Çeliktutan et al. \cite{Çeliktutan2013} & 2013 & 100 & 84.5  & 85  & 72.2  & 80.5  & - \\
\hline
Informative joints based human action recognition using skeleton contexts, Jiang et al. \cite{Jiang2015} & 2015 & 10 & 88.7 & 87.7 & 88.5 & 88.3 & - \\
\hline
Online gesture recognition from pose kernel learning and decision forests, Miranda et al. \cite{Miranda2014} & 2014 & 10 & 96  & 57.1  & 97.3  & 83.5 & - \\
\hline 
Real-Time Gesture Recognition from Depth Data through Key Poses Learning and Decision Forests, Miranda et al. \cite{Miranda2012} & 2012 & 10 & 93.5  & 52  & 95.4  & 80.3  & - \\
\hline 
Space-Time Pose Representation for 3D Human Action Recognition, Devanne et al. \cite{Devanne2013} & 2013 & 10 & 84.8  & 67.8  & 87.1  & 79.9  & - \\
\hline 
\end{tabu} 
\label{tab:miranda-test} 
\end{table} 

Another approach used by some authors is to perform a leave-one-actor-out cross-validation test. In this case, actor invariance is specifically tested by training with all but one actor, and testing the method with the unseen one. This is repeated for all the actors, averaging the returned success rates (see Table~\ref{tab:leave-one-actor-out-test}).

\begin{table}
\caption{Leave-one-actor-out cross-validation test}
\begin{tabu} to \linewidth{|>{\scriptsize}X[5.7,m]|>{\scriptsize}X[1,c]|>{\scriptsize}X[1,c]|>{\scriptsize}X[1,c]|>{\scriptsize}X[1,c]|>{\scriptsize}X[1,c]|>{\scriptsize}X[1,c]|}
\hline
\textbf{Method} & \textbf{Year} & \textbf{AS1} & \textbf{AS2} & \textbf{AS3} & \textbf{Avg.} & \textbf{All} \\
\hline
Evolutionary joint selection to improve human action recognition with RGB-D devices, Chaaraoui et al. \cite{Chaaraoui2014} & 2014 & 91.46  & 91.78  & 97.13  & 93.46  & - \\
\hline 
Fusion of Skeletal and Silhouette-Based Features for Human Action Recognition with RGB-D Devices, Chaaraoui et al. \cite{Chaaraoui2013} & 2013 & 90.65  & 85.15  & 95.93  & 90.58 & - \\
\hline 
3D Action Classification Using Sparse Spatio-temporal Feature Representations, Azary and Savakis \cite{Azary2012} & 2012 & 80.73  & 77.11  & 93.89  & 83.91  & 72.11  \\
\hline 
Grassmannian Sparse Representations and Motion Depth Surfaces for 3D Action Recognition, Azary and Savakis \cite{Azary2013} & 2013 & - & - & - & - & 78.48?\\
\hline 
Fast Exact Hyper-graph Matching with Dynamic Programming for Spatio-temporal Data, \c{C}eliktutan et al. \cite{Çeliktutan2014} & 2014 & - & - & - & - & 72.9?\\
\hline 
\end{tabu} 
\label{tab:leave-one-actor-out-test} 
\end{table}

Finally, in addition to the described validation methods that are frequently used in the literature, there are other authors that have not been included in any table because either the validation method is unclear~\cite{Barnachon2014,Lillo2014} or the employed settings are not used by more than one author~\cite{Ofli2012,Cottone2013,Chaudhry2013,Sabinas2013}. For instance, Ofli et al.~\cite{Ofli2012} use a subset of 17 actions and 8 subjects. They train with 5 subjects and validate with 3 subjects in order to obtain the success rate (41.18\% for the whole dataset). These results are improved in~\cite{Chaudhry2013} with the same setup (83.89\% for the whole dataset). Cottone et al.~\cite{Cottone2013} perform a leave-one-sequence-out cross validation, training with all the sequences in the dataset but one that is used for testing. Then, they perform 10 of these tests obtaining the average success rate (90.47\% for the average of AS1, AS2 and AS3). Sabinas et al.\cite{Sabinas2013} are focused on early detection of gestures, \ie without seeing all the information, and their experimentation is based on one-shot learning. Therefore, their results are not directly comparable (47\% for the average of AS1, AS2 and AS3).

\section{Conclusions}
\label{sec:conclusion}
In this work, we have aimed to give an answer to the question of which is the best action recognition method based on features extracted from depth and skeletal data. Based on the review the present work has performed, it can be observed that we cannot answer this question with confidence. In other words, we cannot know so far. Hence, we have presented the most important divergences in the comparison of action recognition methods that use the MSR Action3D dataset. Among these, we can highlight the mismatch in the number of samples used by most of the works and the different validation methods that have been used. As we have seen, the validation performed by Li et al. is one of the most used. However, the missing information about how to split the dataset into training and validation sets has led to a lot of confusion. Furthermore, most of the authors do not describe how this division is performed in their works. Therefore, experiments cannot be reproduced and fair comparisons cannot be made. Thus, in this work we have tried to clear up the existing confusion. This may enable to improve future comparisons and increase the awareness of the need of clarifying experimental settings.

Among all the validation methods reviewed in this work, we consider that the cross validation considering all the possible splits of the dataset, \ie all the possible combinations (252 tests) of using 5 subjects for training and the remaining ones for testing, is the most robust validation method. However, if testing your method with the 5-5 splits cross validation is very demanding concerning computational cost, then the leave-one-actor-out cross validation is the one we recommend under these conditions.

\section*{Notes to authors}
As it has been difficult in some cases to understand the validation method of the papers, we encourage authors of the reviewed works to contact us in case their works had been misclassified in the previous tables. This way, we will be able to update the document and  correct it. Similarly, authors of new works are also encouraged to contact us in order to incorporate their works if so desired.

\bibliographystyle{splncs}
\bibliography{bibliography}

\begin{thebibliography}{10}

\bibitem{Li2010}
Li, W., Zhang, Z., Liu, Z.:
\newblock {Action recognition based on a bag of 3D points}.
\newblock In: 2010 IEEE Computer Society Conference on Computer Vision and
  Pattern Recognition Workshops (CVPRW). (June 2010)  9--14

\bibitem{Wang2012}
Wang, J., Liu, Z., Wu, Y., Yuan, J.:
\newblock {Mining actionlet ensemble for action recognition with depth
  cameras}.
\newblock In: Computer Vision and Pattern Recognition (CVPR), 2012 IEEE
  Conference on. (June 2012)  1290--1297

\bibitem{Ni2011}
Ni, B., Wang, G., Moulin, P.:
\newblock {RGBD-HuDaAct: A color-depth video database for human daily activity
  recognition}.
\newblock In: 2011 IEEE International Conference on Computer Vision Workshops
  (ICCV Workshops). (Nov 2011)  1147--1153

\bibitem{Jaeyong2012}
Sung, J., Ponce, C., Selman, B., Saxena, A.:
\newblock {Unstructured human activity detection from RGBD images}.
\newblock In: 2012 IEEE International Conference on Robotics and Automation
  (ICRA). (May 2012)  842--849

\bibitem{Xia2012}
Xia, L., Chen, C.C., Aggarwal, J.:
\newblock {View invariant human action recognition using histograms of 3D
  joints}.
\newblock In: 2012 IEEE Computer Society Conference on Computer Vision and
  Pattern Recognition Workshops (CVPRW). (June 2012)  20--27

\bibitem{Fothergill2012}
Fothergill, S., Mentis, H.M., Kohli, P., Nowozin, S.:
\newblock {Instructing people for training gestural interactive systems}.
\newblock In: Proceedings of the SIGCHI Conference on Human Factors in
  Computing Systems, ACM (2012)  1737--1746

\bibitem{Koppula2013}
Koppula, H.S., Gupta, R., Saxena, A.:
\newblock {Learning Human Activities and Object Affordances from RGB-D Videos}.
\newblock International Journal of Robotics Research \textbf{32}(8) (2013)
  951--970

\bibitem{Oreifej2013}
Oreifej, O., Liu, Z.:
\newblock {HON4D: Histogram of Oriented 4D Normals for Activity Recognition
  from Depth Sequences}.
\newblock In: 2013 IEEE Conference on Computer Vision and Pattern Recognition
  (CVPR). (June 2013)  716--723

\bibitem{Kurakin2012}
Kurakin, A., Zhang, Z., Liu, Z.:
\newblock {A real time system for dynamic hand gesture recognition with a depth
  sensor}.
\newblock In: Proceedings of the 20th European Signal Processing Conference
  (EUSIPCO). (Aug 2012)  1975--1979

\bibitem{Wolf2012}
Wolf, C., Mille, J., Lombardi, E., Celiktutan, O., Jiu, M., Baccouche, M.,
  Dellandr{\'e}a, E., Bichot, C.E., Garcia, C., Sankur, B.:
\newblock {The LIRIS Human activities dataset and the ICPR 2012 human
  activities recognition and localization competition}.
\newblock Technical Report RR-LIRIS-2012-004, LIRIS Laboratory (March 2012)

\bibitem{Ofli2013}
Ofli, F., Chaudhry, R., Kurillo, G., Vidal, R., Bajcsy, R.:
\newblock {Berkeley MHAD: A comprehensive Multimodal Human Action Database}.
\newblock In: 2013 IEEE Workshop on Applications of Computer Vision (WACV).
  (Jan 2013)  53--60

\bibitem{Bloom2013}
Bloom, V., Makris, D., Argyriou, V.:
\newblock {G3D: A gaming action dataset and real time action recognition
  evaluation framework}.
\newblock In: 2012 IEEE Computer Society Conference on Computer Vision and
  Pattern Recognition Workshops (CVPRW). (June 2012)  7--12

\bibitem{Cheng2012}
Cheng, Z., Qin, L., Ye, Y., Huang, Q., Tian, Q.:
\newblock {Human Daily Action Analysis with Multi-view and Color-Depth Data}.
\newblock In: Computer Vision - ECCV 2012. Workshops and Demonstrations. Volume
  7584 of Lecture Notes in Computer Science.
\newblock Springer Berlin Heidelberg (2012)  52--61

\bibitem{Theodorakopoulos2014}
Theodorakopoulos, I., Kastaniotis, D., Economou, G., Fotopoulos, S.:
\newblock {Pose-based human action recognition via sparse representation in
  dissimilarity space}.
\newblock Journal of Visual Communication and Image Representation
  \textbf{25}(1) (2014)  12--23

\bibitem{Negin2013}
Negin, F., \"Ozdemir, F., Akg\"ul, C., Y\"uksel, K.A., Er\c{c}il, A.:
\newblock {A Decision Forest Based Feature Selection Framework for Action
  Recognition from RGB-Depth Cameras}.
\newblock In: Image Analysis and Recognition. Volume 7950 of Lecture Notes in
  Computer Science.
\newblock Springer Berlin Heidelberg (2013)  648--657

\bibitem{Munaro2013}
Munaro, M., Ballin, G., Michieletto, S., Menegatti, E.:
\newblock {3D flow estimation for human action recognition from colored point
  clouds}.
\newblock Biologically Inspired Cognitive Architectures \textbf{5} (2013)
  42--51

\bibitem{Seidenari2013}
Seidenari, L., Varano, V., Berretti, S., Del~Bimbo, A., Pala, P.:
\newblock {Recognizing Actions from Depth Cameras as Weakly Aligned Multi-part
  Bag-of-Poses}.
\newblock In: 2013 IEEE Conference on Computer Vision and Pattern Recognition
  Workshops (CVPRW). (June 2013)  479--485

\bibitem{Azary2013}
Azary, S., Savakis, A.:
\newblock {Grassmannian Sparse Representations and Motion Depth Surfaces for 3D
  Action Recognition}.
\newblock In: 2013 IEEE Conference on Computer Vision and Pattern Recognition
  Workshops (CVPRW). (June 2013)  492--499

\bibitem{Gao2014}
Gao, Z., Song, J.m., Zhang, H., Liu, A.A., Xue, Y.b., Xu, G.p.:
\newblock {Human Action Recognition Via Multi-modality Information}.
\newblock Journal of Electrical Engineering \& Technology \textbf{9}(2) (2014)
  739--748

\bibitem{Shen2014}
Shen, X., Zhang, H., Gao, Z., Xue, Y., Xu, G.:
\newblock {Human Behavior Recognition Based on Axonometric Projections and PHOG
  Feature}.
\newblock Journal of Computational Information Systems \textbf{10}(8) (2014)
  3455--3463

\bibitem{Wang2013}
Wang, J., Wu, Y.:
\newblock {Learning Maximum Margin Temporal Warping for Action Recognition}.
\newblock In: 2013 IEEE International Conference on Computer Vision (ICCV).
  (Dec 2013)  2688--2695

\bibitem{Yang2014}
Yang, X., Tian, Y.:
\newblock {Effective 3D action recognition using EigenJoints}.
\newblock Journal of Visual Communication and Image Representation
  \textbf{25}(1) (2014)  2--11

\bibitem{Yuan2014}
Yuan, J., Wang, J., Liu, Z., Wu, Y.:
\newblock {Learning Actionlet Ensemble for 3D Human Action Recognition}.
\newblock IEEE Transactions on Pattern Analysis and Machine Intelligence
  \textbf{36}(5) (2014)  914--927

\bibitem{Wang2015}
Wang, P., Li, W., Gao, Z., Zhang, J., Tang, C., Ogunbona, P.:
\newblock Deep convolutional neural networks for action recognition using depth
  map sequences.
\newblock CoRR \textbf{abs/1501.04686} (2015)

\bibitem{Eweiwi2015}
Eweiwi, A., Cheema, M., Bauckhage, C., Gall, J.:
\newblock Efficient pose-based action recognition.
\newblock In Cremers, D., Reid, I., Saito, H., Yang, M.H., eds.: Computer
  Vision -- ACCV 2014. Volume 9007 of Lecture Notes in Computer Science.
\newblock Springer International Publishing (2015)  428--443

\bibitem{Chen2015}
Chen, C., Jafari, R., Kehtarnava, N.:
\newblock {Action Recognition from Depth Sequences Using Depth Motion
  Maps-based Local Binary Patterns}.
\newblock In: {Proceedings of the IEEE Winter Conference on Applications of
  Computer Vision (WACV)}, Waikoloa Beach, HI (2015)  1092--1099

\bibitem{Rahmani2014eccv}
Rahmani, H., Mahmood, A., Q~Huynh, D., Mian, A.:
\newblock Hopc: Histogram of oriented principal components of 3d pointclouds
  for action recognition.
\newblock In Fleet, D., Pajdla, T., Schiele, B., Tuytelaars, T., eds.: Computer
  Vision – ECCV 2014. Volume 8690 of Lecture Notes in Computer Science.
\newblock Springer International Publishing (2014)  742--757

\bibitem{Chaaraoui2013}
Chaaraoui, A., Padilla-Lopez, J., Florez-Revuelta, F.:
\newblock {Fusion of Skeletal and Silhouette-Based Features for Human Action
  Recognition with RGB-D Devices}.
\newblock In: 2013 IEEE International Conference on Computer Vision Workshops
  (ICCVW). (Dec 2013)  91--97

\bibitem{Chen2013}
Chen, C., Liu, K., Kehtarnavaz, N.:
\newblock {Real-time human action recognition based on depth motion maps}.
\newblock Journal of Real-Time Image Processing (2013)  1--9

\bibitem{Evangelidis2014}
Evangelidis, G., Singh, G., Horaud, R.,  et~al.:
\newblock {Skeletal Quads: Human Action Recognition Using Joint Quadruples}.
\newblock In: Proceedings of the 22nd International Conference on Pattern
  Recognition (ICPR 2014). (2014)

\bibitem{Wang2014}
Wang, J., Liu, Z., Wu, Y.:
\newblock {Random Occupancy Patterns}.
\newblock In: Human Action Recognition with Depth Cameras. Springer Briefs in
  Computer Science.
\newblock Springer International Publishing (2014)  41--55

\bibitem{Luo2013}
Luo, J., Wang, W., Qi, H.:
\newblock {Group Sparsity and Geometry Constrained Dictionary Learning for
  Action Recognition from Depth Maps}.
\newblock In: 2013 IEEE International Conference on Computer Vision (ICCV).
  (Dec 2013)  1809--1816

\bibitem{Zhu2013}
Zhu, Y., Chen, W., Guo, G.:
\newblock {Fusing Spatiotemporal Features and Joints for 3D Action
  Recognition}.
\newblock In: 2013 IEEE Conference on Computer Vision and Pattern Recognition
  Workshops (CVPRW). (June 2013)  486--491

\bibitem{Zou2013}
Zou, W., Wang, B., Zhang, R.:
\newblock {Human Action Recognition by Mining Discriminative Segment with Novel
  Skeleton Joint Feature}.
\newblock In: Advances in Multimedia Information Processing – PCM 2013.
  Volume 8294 of Lecture Notes in Computer Science.
\newblock Springer International Publishing (2013)  517--527

\bibitem{Kapsouras2014}
Kapsouras, I., Nikolaidis, N.:
\newblock {Action recognition on motion capture data using a dynemes and
  forward differences representation}.
\newblock Journal of Visual Communication and Image Representation (2014)  doi:
  10.1016/j.jvcir.2014.04.007

\bibitem{Yang2014b}
Yang, X., Tian, Y.:
\newblock {Super Normal Vector for Activity Recognition Using Depth Sequences}.
\newblock In: {IEEE Conference on Computer Vision and Pattern Recognition
  (CVPR)}. (2014)

\bibitem{Chen2014}
Chen, G., Giuliani, M., Clarke, D., Gaschler, A., Knoll, A.:
\newblock {Action Recognition Using Ensemble Weighted Multi-Instance Learning}.
\newblock In: IEEE International Conference on Robotics and Automation (ICRA).
  (June 2014)

\bibitem{Gowayyed2013}
Gowayyed, M.A., Torki, M., Hussein, M.E., El-Saban, M.:
\newblock {Histogram of Oriented Displacements (HOD): Describing Trajectories
  of Human Joints for Action Recognition}.
\newblock In: Proceedings of the 23rd International Joint Conference on
  Artificial Intelligence. IJCAI'13, AAAI Press (2013)  1351--1357

\bibitem{YangYang2015}
Yang, R., Yang, R.:
\newblock Dmm-pyramid based deep architectures for action recognition with
  depth cameras.
\newblock In Cremers, D., Reid, I., Saito, H., Yang, M.H., eds.: Computer
  Vision -- ACCV 2014. Volume 9007 of Lecture Notes in Computer Science.
\newblock Springer International Publishing (2015)  37--49

\bibitem{Hussein2013}
Hussein, M.E., Torki, M., Gowayyed, M.A., El-Saban, M.:
\newblock {Human Action Recognition Using a Temporal Hierarchy of Covariance
  Descriptors on 3D Joint Locations}.
\newblock In: Proceedings of the 23rd International Joint Conference on
  Artificial Intelligence. IJCAI'13, AAAI Press (2013)  2466--2472

\bibitem{Song2014}
Song, Y., Tang, J., Liu, F., Yan, S.:
\newblock {Body Surface Context: A New Robust Feature for Action Recognition
  From Depth Videos}.
\newblock IEEE Transactions on Circuits and Systems for Video Technology
  \textbf{24}(6) (2014)  952--964

\bibitem{Wang2013b}
Wang, C., Wang, Y., Yuille, A.:
\newblock {An Approach to Pose-Based Action Recognition}.
\newblock In: 2013 IEEE Conference on Computer Vision and Pattern Recognition.
  (June 2013)  915--922

\bibitem{Vieira2014}
Vieira, A.W., Nascimento, E.R., Oliveira, G.L., Liu, Z., Campos, M.F.:
\newblock {On the improvement of human action recognition from depth map
  sequences using Space–Time Occupancy Patterns}.
\newblock Pattern Recognition Letters \textbf{36}(15) (2014)  221--227

\bibitem{Vieira2012}
Vieira, A., Nascimento, E., Oliveira, G., Liu, Z., Campos, M.:
\newblock {STOP: Space-Time Occupancy Patterns for 3D Action Recognition from
  Depth Map Sequences}.
\newblock In: Progress in Pattern Recognition, Image Analysis, Computer Vision,
  and Applications. Volume 7441 of Lecture Notes in Computer Science.
\newblock Springer Berlin Heidelberg (2012)  252--259

\bibitem{Wang2013c}
Wang, R., Medioni, G., Winstein, C., Blanco, C.:
\newblock {Home Monitoring Musculo-skeletal Disorders with a Single 3D Sensor}.
\newblock In: 2013 IEEE Conference on Computer Vision and Pattern Recognition
  Workshops (CVPRW). (June 2013)  521--528

\bibitem{Zhao2013}
Zhao, X., Li, X., Pang, C., Zhu, X., Sheng, Q.Z.:
\newblock Online human gesture recognition from motion data streams.
\newblock In: Proceedings of the 21st ACM International Conference on
  Multimedia. MM '13, New York, NY, USA, ACM (2013)  23--32

\bibitem{Li2013}
Li, X., Sheng, Q., Pang, C., Zhao, X., Wang, S.:
\newblock {Effective approaches in human action recognition}.
\newblock In: 2013 International Conference on Advanced Computer Science and
  Information Systems (ICACSIS),. (Sept 2013)  1--7

\bibitem{Qin2013}
Qin, S., Yang, Y., Jiang, Y.:
\newblock {Gesture recognition from depth images using motion and shape
  features}.
\newblock In: 2nd International Symposium on Instrumentation and Measurement,
  Sensor Network and Automation (IMSNA). (Dec 2013)  172--175

\bibitem{Althloothi2014}
Althloothi, S., Mahoor, M.H., Zhang, X., Voyles, R.M.:
\newblock {Human activity recognition using multi-features and multiple kernel
  learning}.
\newblock Pattern Recognition \textbf{47}(5) (2014)  1800--1812

\bibitem{Liang2013}
Liang, B., Zheng, L.:
\newblock {Three Dimensional Motion Trail Model for Gesture Recognition}.
\newblock In: 2013 IEEE International Conference on Computer Vision Workshops
  (ICCVW). (Dec 2013)  684--691

\bibitem{Venkataraman2013}
Venkataraman, V., Turaga, P., Lehrer, N., Baran, M., Rikakis, T., Wolf, S.:
\newblock {Attractor-Shape for Dynamical Analysis of Human Movement:
  Applications in Stroke Rehabilitation and Action Recognition}.
\newblock In: 2013 IEEE Conference on Computer Vision and Pattern Recognition
  Workshops (CVPRW). (June 2013)  514--520

\bibitem{Ellis2013}
Ellis, C., Masood, S., Tappen, M., LaViola, J., Sukthankar, R.:
\newblock {Exploring the Trade-off Between Accuracy and Observational Latency
  in Action Recognition}.
\newblock International Journal of Computer Vision \textbf{101}(3) (2013)
  420--436

\bibitem{Wang2012b}
Wang, J., Liu, Z., Chorowski, J., Chen, Z., Wu, Y.:
\newblock {Robust 3D Action Recognition with Random Occupancy Patterns}.
\newblock In: Proceedings of the 12th European conference on Computer
  Vision-Volume Part II.
\newblock Springer Berlin Heidelberg (2012)  872--885

\bibitem{Tran2013b}
Tran, Q., Ly, N.:
\newblock {Sparse spatio-temporal representation of joint shape-motion cues for
  human action recognition in depth sequences}.
\newblock In: 2013 IEEE International Conference on Computing and Communication
  Technologies, Research, Innovation, and Vision for the Future (RIVF). (Nov
  2013)  253--258

\bibitem{Zanfir2013}
Zanfir, M., Leordeanu, M., Sminchisescu, C.:
\newblock {The Moving Pose: An Efficient 3D Kinematics Descriptor for
  Low-Latency Action Recognition and Detection}.
\newblock In: 2013 IEEE International Conference on Computer Vision (ICCV).
  (Dec 2013)  2752--2759

\bibitem{Tran2013}
Tran, Q., Ly, N.:
\newblock {An effective fusion scheme of spatio-temporal features for human
  action recognition in RGB-D video}.
\newblock In: 2013 International Conference on Control, Automation and
  Information Sciences (ICCAIS). (Nov 2013)  246--251

\bibitem{Rahmani2014wacv}
Rahmani, H., Mahmood, A., Mian, A., Huynh, D.:
\newblock Real time action recognition using histograms of depth gradients and
  random decision forests.
\newblock In: Proceedings of the IEEE Winter Applications of Computer Vision
  Conference (WACV). (2014)

\bibitem{Soh2012}
Soh, H., Demiris, Y.:
\newblock {Iterative temporal learning and prediction with the sparse online
  echo state gaussian process}.
\newblock In: 2012 International Joint Conference on Neural Networks (IJCNN).
  (June 2012)  1--8

\bibitem{Ohn-Bar2013}
Ohn-Bar, E., Trivedi, M.:
\newblock {Joint Angles Similarities and HOG2 for Action Recognition}.
\newblock In: 2013 IEEE Conference on Computer Vision and Pattern Recognition
  Workshops (CVPRW). (June 2013)  465--470

\bibitem{Luo2014}
Luo, J., Wang, W., Qi, H.:
\newblock {Spatio-temporal feature extraction and representation for RGB-D
  human action recognition}.
\newblock Pattern Recognition Letters (2014)  doi: 10.1016/j.patrec.2014.03.024

\bibitem{Rahmani2014icpr}
Rahmani, H., Mahmood, A., Huynh, D., Mian, A.:
\newblock Action classification with locality-constrained linear coding.
\newblock In: Proceedings of the 22nd International Conference on Pattern
  Recognition (ICPR 2014). (2014)

\bibitem{Xia2013}
Xia, L., Aggarwal, J.:
\newblock {Spatio-temporal Depth Cuboid Similarity Feature for Activity
  Recognition Using Depth Camera}.
\newblock In: 2013 IEEE Conference on Computer Vision and Pattern Recognition
  (CVPR),. (June 2013)  2834--2841

\bibitem{Climent2013}
Climent-Perez, P., Chaaraoui, A.A., Padilla-Lopez, J.R., Florez-Revuelta, F.:
\newblock {Optimal Joint Selection for Skeletal Data from RGB-D Devices Using a
  Genetic Algorithm}.
\newblock In: Advances in Computational Intelligence. Volume 7630 of Lecture
  Notes in Computer Science.
\newblock Springer Berlin Heidelberg (2013)  163--174

\bibitem{Miranda2012}
Miranda, L., Vieira, T., Martinez, D., Lewiner, T., Vieira, A., Campos, M.:
\newblock {Real-Time Gesture Recognition from Depth Data through Key Poses
  Learning and Decision Forests}.
\newblock In: 25th Conference on Graphics, Patterns and Images (SIBGRAPI). (Aug
  2012)  268--275

\bibitem{Çeliktutan2013}
Celiktutan, O., Akgul, C.B., Wolf, C., Sankur, B.:
\newblock {Graph-based Analysis of Physical Exercise Actions}.
\newblock In: Proceedings of the 1st ACM International Workshop on Multimedia
  Indexing and Information Retrieval for Healthcare. MIIRH '13, New York, NY,
  USA, ACM (2013)  23--32

\bibitem{Çeliktutan2014}
\c{C}eliktutan, O., Wolf, C., Sankur, B., Lombardi, E.:
\newblock {Fast Exact Hyper-graph Matching with Dynamic Programming for
  Spatio-temporal Data}.
\newblock Journal of Mathematical Imaging and Vision (March 2014)  1--21

\bibitem{Chaaraoui2014}
Chaaraoui, A.A., Padilla-Lepez, J.R., Climent-Perez, P., Florez-Revuelta, F.:
\newblock {Evolutionary joint selection to improve human action recognition
  with RGB-D devices}.
\newblock Expert Systems with Applications \textbf{41}(3) (2014)  786 -- 794

\bibitem{ChenG2015}
Chen, G., Clarke, D., Giuliani, M., Gaschler, A., Knoll, A.:
\newblock Combining unsupervised learning and discrimination for 3d action
  recognition.
\newblock Signal Processing \textbf{110}(0) (2015)  67 -- 81 Machine learning
  and signal processing for human pose recovery and behavior analysis.

\bibitem{Azary2012}
Azary, S., Savakis, A.:
\newblock {3D Action Classification Using Sparse Spatio-temporal Feature
  Representations}.
\newblock In: Advances in Visual Computing. Volume 7432 of Lecture Notes in
  Computer Science.
\newblock Springer Berlin Heidelberg (2012)  166--175

\bibitem{Jiang2015}
Jiang, M., Kong, J., Bebis, G., Huo, H.:
\newblock Informative joints based human action recognition using skeleton
  contexts.
\newblock Signal Processing: Image Communication \textbf{33}(0) (2015)  29 --
  40

\bibitem{Miranda2014}
Miranda, L., Vieira, T., Martínez, D., Lewiner, T., Vieira, A.W., Campos,
  M.F.M.:
\newblock {Online gesture recognition from pose kernel learning and decision
  forests}.
\newblock Pattern Recognition Letters \textbf{39} (2014)  65--73

\bibitem{Devanne2013}
Devanne, M., Wannous, H., Berretti, S., Pala, P., Daoudi, M., Del~Bimbo, A.:
\newblock {Space-Time Pose Representation for 3D Human Action Recognition}.
\newblock In: New Trends in Image Analysis and Processing – ICIAP 2013.
  Volume 8158 of Lecture Notes in Computer Science.
\newblock Springer Berlin Heidelberg (2013)  456--464

\bibitem{Barnachon2014}
Barnachon, M., Bouakaz, S., Boufama, B., Guillou, E.:
\newblock {Ongoing human action recognition with motion capture}.
\newblock Pattern Recognition \textbf{47}(1) (2014)  238--247

\bibitem{Lillo2014}
Lillo, I., Niebles, J., Soto, A.:
\newblock {Discriminative Hierarchical Modeling of Spatio-Temporally Composable
  Human Activities}.
\newblock In: 2014 IEEE Conference on Computer Vision and Pattern Recognition
  (CVPR). (2014)

\bibitem{Ofli2012}
Ofli, F., Chaudhry, R., Kurillo, G., Vidal, R., Bajcsy, R.:
\newblock {Sequence of the Most Informative Joints (SMIJ): A new representation
  for human skeletal action recognition}.
\newblock In: 2012 IEEE Computer Society Conference on Computer Vision and
  Pattern Recognition Workshops (CVPRW). (June 2012)  8--13

\bibitem{Cottone2013}
Cottone, P., Re, G., Maida, G., Morana, M.:
\newblock {Motion sensors for activity recognition in an ambient-intelligence
  scenario}.
\newblock In: 2013 IEEE International Conference on Pervasive Computing and
  Communications Workshops (PERCOM Workshops). (March 2013)  646--651

\bibitem{Chaudhry2013}
Chaudhry, R., Ofli, F., Kurillo, G., Bajcsy, R., Vidal, R.:
\newblock {Bio-inspired Dynamic 3D Discriminative Skeletal Features for Human
  Action Recognition}.
\newblock In: 2013 IEEE Conference on Computer Vision and Pattern Recognition
  Workshops (CVPRW). (June 2013)  471--478

\bibitem{Sabinas2013}
Sabinas, Y., Morales, E., Escalante, H.:
\newblock {A One-Shot DTW-Based Method for Early Gesture Recognition}.
\newblock In: Progress in Pattern Recognition, Image Analysis, Computer Vision,
  and Applications. Volume 8259 of Lecture Notes in Computer Science.
\newblock Springer Berlin Heidelberg (2013)  439--446

\end{thebibliography}
\end{document}